\documentclass[10pt,twocolumn,letterpaper]{article}

\usepackage{cvpr}
\usepackage{times}
\usepackage{epsfig}
\usepackage{graphicx}
\usepackage{amsmath}
\usepackage{amssymb}
\usepackage{soul,color}
\usepackage{amsmath}
\usepackage[ruled,lined]{algorithm2e}
\usepackage{tabularx}
\usepackage{booktabs}
\usepackage{epstopdf}
\usepackage{color, colortbl}
\usepackage{makecell}
\usepackage{multirow}
\usepackage{caption}

\usepackage[pagebackref=true,breaklinks=true,letterpaper=true,colorlinks,bookmarks=false]{hyperref}

\cvprfinalcopy 

\def\cvprPaperID{9726} 

\begin{document}

\title{Tackling the Problem of Limited Data and Annotations in Semantic Segmentation}

\author{Ahmadreza Jeddi \\
David R. Cheriton School of Computer Science, University of Waterloo, Waterloo, Ontario, Canada\\
a2jeddi@uwaterloo.ca 
}

\maketitle
\thispagestyle{empty}

\begin{abstract}
In this work, the case of semantic segmentation on a small image dataset (simulated by 1000 randomly selected images from PASCAL VOC 2012), where only weak supervision signals (scribbles from user interaction) are available is studied. Especially, to tackle the problem of limited data annotations in image segmentation, transferring different pre-trained models and CRF based methods are applied to enhance the segmentation performance. To this end, RotNet, DeeperCluster, and Semi\&Weakly Supervised Learning (SWSL) pre-trained models are transferred and finetuned in a DeepLab-v2 baseline, and dense CRF is applied both as a post-processing and loss regularization technique. The results of my study show that, on this small dataset, using a pre-trained ResNet50 SWSL model gives results that are 7.4\% better than applying an ImageNet pre-trained model; moreover, for the case of training on the full PASCAL VOC 2012 training data, this pre-training approach increases the mIoU results by almost 4\%. On the other hand, dense CRF is shown to be very effective as well, enhancing the results both as a loss regularization technique in weakly supervised training and as a post-processing tool. All the codes are publicly available at  \url{https://github.com/Ahmadreza-Jeddi/rloss}
\end{abstract}

\vspace{-0.2cm}
\section{Introduction}
\vspace{-0.2cm}
Models based on CNN architectures provide the state-of-the-art performances on the well-known semantic segmentation datasets such as Pascal VOC 2012~\cite{everingham2015pascal}, Microsoft COCO~\cite{lin2014microsoft}, and Cityscapes~\cite{cordts2016cityscapes}. Especially, algorithms based on FCN~\cite{long2015fully} and Deeplab~\cite{chen2017deeplab} and their variations reach almost human level performances on these tasks. However, we need to consider that these tasks enjoy full pixel-level annotations which leads them into their superior outcomes. Compared to other computer vision tasks like object recognition and object detection, generating fully supervised annotations for semantic segmentation is much more expensive, and in many cases, even the provided ground truth may be prone to errors caused by human experts.

Numerous studies have tried to provide solutions to tackle the problem of costly supervision in segmentation; weak and semi supervised approaches~\cite{papandreou2015weakly, tang2018regularized, pathak2015constrained}, as well as CRF based methods~\cite{tang2018regularized, chen2017deeplab, kolesnikov2016seed} are examples of these studies. In this work, I study the solutions that help boost the performance of the segmentation models in scenarios where the amount of the available training data is small and only weak supervisions in the form of user interactions are available. The small training data assumption is a realistic one for many problems such as medical imaging. As for the weak interactive supervision, a huge body of research~\cite{rother2004grabcut, boykov2001interactive, lin2016scribblesup, mortensen1999toboggan} has been done in this field, where small weak annotations such as few points, boxes, and scribbles are provided by the user.

In the past few years, a very popular line of work involving CNNs has been pre-training CNN models on large scales of data and then fine-tuning these models for many downstream tasks. This approach is adopted in almost every industrial and academic project to increase both the performance and efficiency of the models. Supervised models trained on the ImageNet~\cite{deng2009imagenet} dataset have been central for many transfer learning methods, and have enabled state-of-the-art results on tasks such as object detection~\cite{ren2015faster} and semantic segmentation~\cite{chen2017deeplab}. Meanwhile, many semi-supervised, weakly-supervised, and self-supervised techniques have been proposed in the literature to learn feature representations with zero or very small expert supervision. By increasing the size of the training samples using unlabeled data, using only weak supervision signals, and defining unsupervised pretext tasks, these methods have been able to learn representations that have been quite effective to boost the performance of the downstream tasks. However, most of them have been transferred to classification and detection problems; in this work, I will transfer and fine-tune these models for semantic segmentation models.

\begin{figure}
\vspace{-0.8cm}
    \centering
    \includegraphics[width=8cm]{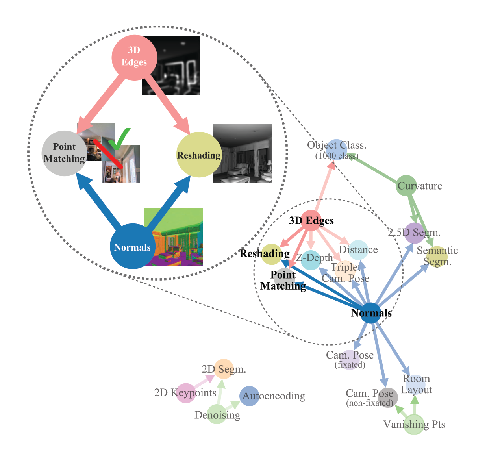}
    \vspace{-0.25cm}
    \caption{Transferability relationships among various computer vision tasks~\cite{zamir2018taskonomy}}
    \vspace{-0.15cm}
    \label{fig:taskonomy}
\end{figure}

In the recent work \cite{zamir2018taskonomy}, Zamir \textit{et al.} provide a thorough study of the space of transferability of the visual tasks, and show the transferability relationships among 26 different visual tasks in multiple visual dimensions (2D, 2.5D, and 3D). Figure \ref{fig:taskonomy} shows these relationships; they use transfer learning dependencies among those 26 visual tasks to extract the non-trivial relationships shown in the figure. As can be seen in Figure \ref{fig:taskonomy}, models trained for finding surface normals or models trained for extracting image curvatures can successfully transfer to semantic segmentation tasks. On the other hand, these findings suggest that models trained for image classification may not be very good pre-training options for semantic segmentation. This study further indicates the difference between image classifiers/detectors and image segmentation models; the former only require abstract representations for their success, while the latter need to focus on fine-grained details. Nonetheless, various studies~\cite{chen2017deeplab, tang2018regularized, long2015fully} have shown that using pre-training can help both the performance and the convergence speed of the semantic segmentation, compared to the case of randomly initializing the network parameters. While most of these pre-trained models have been fully supervised ImageNet classifiers, here, I intend to transfer the most recent semi-supervised and self-supervised models trained on different types of visual datasets. More explanations on these methods and experimental results is provided in sections \ref{sec:method} and \ref{sec:experiment}.

Another class of segmentation methods that can be effective for my case of study (small and weakly-supervised dataset), are non deep learning based ones\footnote{Tang \textit{et al.} \cite{tang2018regularized} use the term \textit{shallow} segmentation in order to refer to these techniques; however, since \textit{shallow} might be mistaken as a technique opposed to deep learning, I avoid using it, and instead, I call these methods \textit{non DL-based methods}, techniques like graph-cut, random walker, and etc.}; before the deep learning era, these methods were the \textit{de facto} tools for segmentation tasks. They work best when there are (weak) supervisions available for the image. A very useful feature of these methods is that unlike CNN based models, they do not need a large dataset to learn from; instead, they work for each sample independently. For samples with weak supervisions, they can provide sufficiently accurate segmentation masks.

In recent years, several works have adopted the non DL-based techniques to the context of CNN based segmentation. Methods like \cite{rajchl2016deepcut, lin2016scribblesup, papandreou2015weakly} use the weak supervision and the non DL based optimization techniques as a pre-processing step to generate fully-labeled masks (proposals), and afterward, they treat these proposals as ground truth labels and imitate the full-supervised semantic segmentation. CRFs as a popular class of non DL-based methods have been used for post-processing the results of CNN models~\cite{chen2017deeplab}, and as trainable layers~\cite{zheng2015conditional} as well. And recently, the seminal works of Tang \textit{et al.} \cite{tang2018regularized, tang2018normalized} propose to add the non DL-based techniques to the neural network as regularized losses. The flexibility of the gradient descent allows them to model dense CRF, normalized cut, and their combination, kernel cut, as regularization losses which can be added to the (partial) cross-entropy loss and the combination uses the gradient descent as the optimizer. They provide exhaustive experiments on the famous Pascal VOC 2012 segmentation problem; the results indicate that the regularized losses are a promising tool for weakly supervised segmentation, and I will apply these loss regularization techniques in this study.

Therefore, to put everything together, I am studying the effects of transfer learning and loss regularization techniques on scenarios where the size of training data is small, and the only supervision comes from the interactive inputs from the user, in which they merely highlight the regions of interest in the images (scribbles). To this end, my contributions can be folded as below:

\begin{itemize}
    \item The case of weak supervision for semantic segmentation is studied for a small dataset; in this work, I choose Pascal VOC 2012, and from this dataset, I randomly select 1000 images.
    \item The effects of two completely different sets of methods are studied for the case of limited data size; transfer learning, and loss regularization techniques (in this work, dense CRF). The effects of both sets of methods have been separately studied in previous works, and on different scales, they have been shown to be effective techniques. In this work, I combine these methods and focus on the case of small data size.
    \item Exhaustive experiments using the DeepLab-v2 as the baseline are conducted and the reported results further indicate the effectiveness of the two sets of techniques. Especially, I observe that transferring and fine-tuning models pre-trained by SWSL~\cite{yalniz2019billion} can increase the mIoU results on the PASCAL VOC 2012 \textit{val} set by 7\% compared to those of training with a model pre-trained on ImageNet.
\end{itemize}

The rest of this paper is organized as follows: section \ref{sec:bg} discusses the related topics and gives a brief description for each one. Section \ref{sec:method} describes the pipeline of the semantic segmentation task and the specific components that I have chosen for this study, after which, experimental results are reported in section \ref{sec:experiment}. Finally, the work is concluded in section \ref{sec:conclusion}.

\section{Background}
\label{sec:bg}
In this section, I will describe the related topics in more details.

\subsection{Transfer Learning}
Transfer learning is the task of applying a model trained on a large body of data to ones with smaller labeled or unlabeled data. In most cases, the transferred models are classification models learned on large-scale datasets. Depending on the problem, all or parts of the transferred model might be used, and the target task can choose to change the parameters of the representation learning layers of the source model (fine-tuning) or freeze those layers and just modify the final classifier layers (assuming the source model is a classifier). In academia and commercial uses, it is common to transfer classification models into object detection or semantic segmentation. These target tasks usually consist of several components, one of which is called the encoder, which embeds the input space into a latent representation space that can be folded to form a manifold of the input space. During the transfer learning, the feature extraction part of the source classifier replaces the encoder of the target tasks, and commonly, the fine-tuning is performed.

\subsection{Self-Supervised Learning}
Self-supervised learning is a subclass of unsupervised learning, where in order to learn visual representations, one defines a pretext task based on the heuristic algorithms that rely on the inherent features of images and videos. The pretext task is usually a classification problem, and ideally, solving this task requires understanding some visual semantics. The models utilized to solve the problem can learn and measure their performances using the common full supervision objective functions (e.g., cross-entropy); however, the labels are automatically obtained from the pretext task. In the past few years, a huge number of self-supervised models have been proposed, with \cite{gidaris2018unsupervised, caron2018deep, caron2019unsupervised, oord2018representation, noroozi2018boosting, kolesnikov2019revisiting, feng2019self, doersch2017multi} as some of the more recent ones. Solving jigsaw puzzles made from image patches~\cite{noroozi2016unsupervised}, rotating an image by a multiple of 90 degrees and predicting the rotation~\cite{gidaris2018unsupervised}, predicting the relative position of an image patch to the central patch~\cite{doersch2015unsupervised} are examples of the pretext tasks that are defined to add the self-supervision. Since the pretext tasks do not usually have much practical use on their own, self-supervised models are almost always transferred to other (downstream) tasks to boost the initial state of the optimization problem, and then fine-tuned to tackle the new task.

\subsection{Semi-Supervised Learning}
It is a learning paradigm that considers cases where both labeled and unlabeled data are available. Its goal is to find ways of using large scales of available unlabeled data to improve the performance of supervised methods where their annotated data is scarce and it is costly to get labeled data~\cite{zhu2009introduction}. In the context of machine learning, this paradigm is mostly used for classification problems, where over the years, a huge body of research has tried to increase the size of datasets by label propagation from labeled data to unlabeled data, and hopefully, improve the classification. Applying variational methods through approximate Bayesian inference and deep generative models~\cite{kingma2014semi}, and Generative Adversarial Networks (GANs)~\cite{odena2016semi} are examples of some more recent DL based label propagation methods, which have shown tremendous success. To find more examples of the works done on semi-supervised learning, the reader can refer to \cite{van2020survey}.

\subsection{DeepLab}
State-of-the-art CNN-based semantic segmentation models are usually derivatives of two well-known models in this field; Fully Convolutional Networks (FCNs)~\cite{long2015fully} and DeepLab~\cite{chen2017deeplab}. These two methods share almost the same structure, in which at first the input image's features are extracted by an encoder, and then the extracted features are upsampled to generate the corresponding segmentation mask. The FCN models use deconvolution layers, so they train the upsampling component of the segmentation. On the other hand, DeepLab simply uses bilinear interpolation for the upsampling purpose. However, using bilinear interpolation does not mean that DeepLab is ineffective; through its multiple versions and refinements, DeepLab now manages to increase the resolution of the intermediate features by removing the max-pooling and down-sampling (striding) components, and introduces atrous convolutions~\cite{holschneider1990real, papandreou2015modeling} which have the ability of enlarging the field of view for the convolution kernels. Atrous convolutions increase feature map resolutions which is key for semantic segmentation. The bilinear interpolation is then applied to retrieve the original image size. This is what DeepLab v2 does; there are newer versions of DeepLab which are far more powerful, but for the purposes of this study, I choose DeepLab v2 as the baseline.

\subsection{Weakly Supervised Semantic Segmentation}
\begin{figure*}
\vspace{-0.8cm}
    \centering
    \includegraphics[width=18cm]{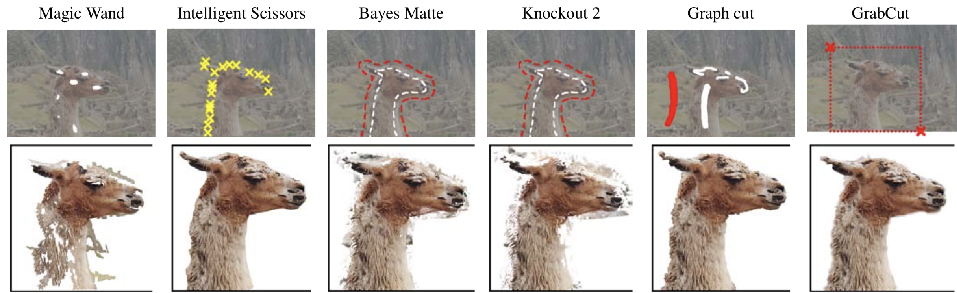}
    \caption{Top row images show the user interaction required for each segmentation method. And the bottom row images show the resulting segmentation masks~\cite{rother2004grabcut}.}
    \vspace{0.40cm}
    \label{fig:interactive}
\end{figure*}

\begin{figure*}
\vspace{-0.8cm}
    \centering
    \includegraphics[width=18cm]{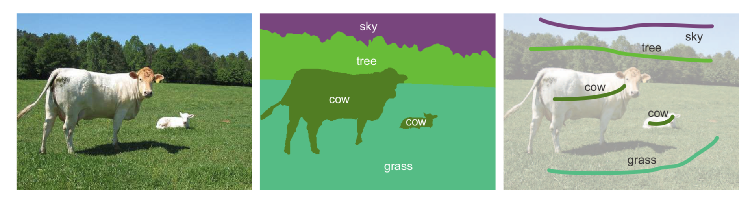}
    \caption{From left to right: image, ground truth masks, and interactive user scribbles on each object~\cite{lin2016scribblesup}.}
    \vspace{-0.30cm}
    \label{fig:scribble}
\end{figure*}

Yet another learning paradigm, where the supervision signals come from limited sources. In \cite{han2014object, bilen2016weakly}, the model uses only the image-level tags (indicating which objects are in the image) to train object detection models. \cite{pathak2015constrained} uses image level tags and Constrained CNN to perform semantic segmentation. These are some examples of using weak signals to train powerful predictive models. Although it is well understood that these supervisions are not perfect, they have been widely used to alleviate the need for large hand-labeled datasets\cite{zhou2018brief}, especially when data labelling is costly (in terms of time and/or expense). The manner of the weak supervision mainly depends on the type of vision task at hand and the available supervision cues. This paradigm has been especially popular in semantic segmentation, where compared to other visual tasks, full supervision is much more expensive to achieve, additionally, these annotations may be prone to subjective errors during hand labeling. Interactive supervision has been used as a type of weak supervision in semantic segmentation for a long time. Figure \ref{fig:interactive} shows some of the examples of this type of weak supervision and the corresponding segmentation method that uses each one. Creating scribbles by dragging the cursor in the center of the target objects is another way of providing interactive weak supervisions. Figure \ref{fig:scribble} shows an example of such weak signals from the ScribbleSuP~\cite{lin2016scribblesup} annotations on the PASCAL VOC 2012 dataset. From here on, when I refer to weak annotations, I mean the scribbles on an image, which come from the ScribbleSup dataset. 

\subsection{Non DL-based Methods for Unsupervised Segmentation}
\label{subsec:nondl}
Deep CNN models are data-hungry; they can achieve human-level (and sometimes better) performance if supplied with enough data (and supervision). But, what about a single image? Consider a case where there are no training data and we only have one single image and we may or may not have some weak supervision available; how can we find the semantic segmentation in this scenario? There is another class of methods that have been around for much longer than the modern CNN based methods; they are the non DL-based methods. These methods usually formulate the segmentation problem as an energy minimization based on two visual cues: RGB colors and edges in the image. Boundary-based methods such as intelligent scissors~\cite{mortensen1999toboggan}, and active contours~\cite{kass1988snakes} focus on image edges, and region-based methods such as thresholding~\cite{tobias2002image} and region growing~\cite{adams1994seeded} focus on fitting models on the pixel features, RGB and/or XY (location), of the objects of interest and separating them from the background (other objects of the scene).

There is a third class, which consists of those methods combining both region and boundary energies.
CRF-based techniques (e.g. graph cuts)~\cite{boykov2003computing}, and random walker~\cite{grady2004multi} are examples of this category. Especially, formulating the segmentation task as a CRF model has been very popular in the literature; a typical technique of this family consists of two components: the unary and the pairwise potentials. The former potentials model the local information about each pixel, e.g., which class the pixel belongs (region-based). The latter enforce boundary consistencies, e.g., how different the pixels of two neighboring segments are (boundary-based). CRF technique has been adopted to the modern CNN-based segmentation models as well. They can be added to CNN segmentation models in 3 different ways: the simplest approach is to just apply the CRF model to the output mask of the CNN model (post-processing), the second case applies it as trainable layers, and the last one that has been recently proposed by Tang \textit{et al.}~\cite{tang2018regularized}, is to incorporate it with the CNN model's loss function as a regularization term.

\vspace{-0.25cm}
\section{Methodology}
\label{sec:method}
In this section, I will explain the solutions that are applied to overcome the problem of semantic segmentation on a small weakly supervised dataset. To this end, pre-trained models that are transferred and fine-tuned and the non DL-based segmentation techniques that are applied to enhance the segmentation quality are described. A very important fact about the pre-trained methods and fine-tuning is that, since these models are trained on very large scales of data (hundreds of millions), it is not possible for me to reproduce the models or customize them; as a result of this fact, I use the pre-trained models as they are and only fine-tune them during the DeepLab training. Additionally, from the available pre-trained models, I select those whose baseline architectures are ResNet50 or VGG16.

\subsection{RotNet}
In \cite{gidaris2018unsupervised}, Gidaris \textit{et al.} propose RotNet. It is one of the best self-supervised methods to learn visual representations. RotNet defines a pretext task, in which input images are randomly rotated by one of the 4 possible cases in $\{ 0^\circ, 90^\circ, 180^\circ, 270^\circ \}$, and the network predicts the rotation. In the original work, Gidaris \textit{et al.} train the pretext task on the ImageNet dataset using AlexNet as the baseline. However, very recently, Caron \textit{et al}~\cite{caron2019unsupervised} trained a RotNet on YFCC100M~\cite{thomee2016yfcc100m} using a VGG16 model. In this work, I will use this pre-trained VGG16 model for the model transfer purposes, especially, its feature extractor part will be used as the encoder in the DeepLab-v2 architecture.

\subsection{DeeperCluster}
This self-supervised method combines RotNet with the deep clustering method proposed in \cite{caron2018deep}. In \cite{caron2018deep}, Caron \textit{et al.} first use k-means to cluster the input images into \textit{k} different clusters (based on the CNN extracted features), this clustering forms pseudo-labels, and then a classifier is trained upon the learned features, which assigns each input to one of the clusters. So there are two sets of parameters, one is the set of CNN parameters that learn to create the representation of input images, and the other is the set of the classifier parameters which assign input features to clusters. the deep clustering method alternates between learning these two sets of parameters, updating one when the other set is fixed, and after every few epochs, the k-means algorithm is re-run and pseudo-labels are updated. DeeperCluster~\cite{caron2019unsupervised} combines deep clustering and RotNet, so the model learns two pretext tasks together. Their feature learning component is a VGG16 model trained on YFCC100, which I will use it as the encoder of the DeepLab and fine-tune it.

\subsection{Semi-Weakly Supervised ImageNet Model}
In \cite{yalniz2019billion}, Yalniz \textit{et al.} train an ImageNet classifier using a teacher-student method as follows:  first, a powerful ImageNet classifier is trained as the teacher model; then, this teacher model is applied on a very large unlabeled dataset (YFCC100M), and top K samples from each ImageNet class from the results of teacher classification are extracted. A new student model is then trained on the dataset created by semi-supervision, and finally, this student model is fine-tuned on the original ImageNet dataset itself. The teacher model itself is trained using weak supervision signals, in which, weakly annotated data by hashtags from the IG-1B-Targeted~\cite{mahajan2018exploring} dataset are selected for each ImageNet class, and then the teacher is first pre-trained on this dataset, and then fine-tuned on ImageNet itself. I will refer to this model as SWSL (Semi-Weakly Supervised Learning). The ResNet50 implementation of this approach reaches the top-5 accuracy of 94.9\% on ImageNet, and I select this model for the purpose of DeepLab pre-training.

\subsection{CRF based segmentation}
As briefly mentioned in \ref{subsec:nondl}, CRF/MRF probabilistic graphical models are a class of non DL-based techniques widely used in semantic segmentation. In these approaches, images are modeled as graphs, where pixels are the nodes and edges exist between pixels for which a pairwise potential is defined (higher-order modeling such as super-pixels~\cite{fulkerson2009class} is possible as well, but I will only consider the simple case of pixel-level graphs). The probabilistic graphical models assume pixels as random variables and based on the prior knowledge that might be known about these random variables, they focus on inferring the latent ground truth distribution which corresponds to the ideal image segmentation.

CRFs model the segmentation problem as a label assignment task which is formulated as the following energy minimization:
\begin{align}
    E(S, I) = \sum_{p \in I}^{} \phi_l (T_p = S_p | I) \hspace{1em} + \hspace{3em} \nonumber \\ \hspace{3em} \sum_{p,q \in I}^{} \psi_{p,q} (T_p = S_p, T_q = S_q | I)
    \label{eq:crf_energy}
\end{align}
, where \textit{I} shows the image, \textit{S} a given segmentation, and \textit{T} shows the ground truth label of each pixel. The first part of the energy equation shows the unary potentials for each class ($\phi$), and the second part models the pairwise potentials ($\psi$). CRFs are extremely popular and different versions of them exist in the literature, like grid CRF which only considers edges between neighboring pixels, and dense CRF in which every two pixels have a pairwise potential defined for them. The former still suffers the locality curse, and the latter is very computationally expensive to perform. Although the the minimization of the energy shown in equation \ref{eq:crf_energy} is NP-hard, there are few methods that can solve it exactly or approximately; especially, for the dense CRF case, if the pairwise potentials are Gaussian, the minimization becomes a bilateral filtering problem~\cite{paris2006fast, krahenbuhl2011efficient} for which many fast solvers exist.

In recent years, CRFs have been used along almost every CNN-based segmentation model. Some use it for post-processing~\cite{chen2017deeplab, krahenbuhl2011efficient}, some as appended trainable layers~\cite{zheng2015conditional, barron2016fast}. The most recent approach is the work done by Tang \textit{et al.}; In \cite{tang2018regularized}, they incorporate the dense CRF as a regularization loss with the well-known cross-entropy loss, and the combination of these two losses is then optimized using the gradient descent. When applying this technique on the weakly supervised training data, they apply the cross-entropy loss only on the partially known labels, which they call it partial cross-entropy (pCE). They experimentally show that using only the pCE loss results in an acceptable performance, and when they combine the two losses they achieve near full supervision performance. For extending the pCE loss with the regularized CRF one, the unary potentials are first derived from the CNN (the softmax outputs), then, the pairwise dense CRF energy is calculated based on the softmax soft labels, and then, the corresponding loss is merged with the pCE one to be optimized. In this work, I will mainly focus on this final version of using CRF; applying it as a regularization loss. In some experiments, I will use the CRF post-processing as well. Results show that both approaches enhance the segmentation performance.

\section{Experiments}
\label{sec:experiment}
In this section, after discussing the experimental setup, I will illustrate the effects of the different techniques that are applied to tackle the problem of semantic segmentation on a small and partially annotated dataset.

\subsection{Experimental Setup}
\textbf{Dataset:} All experiments are done on the PASCAL VOC 2012 segmentation dataset. Especially, in order to simulate the case of a small and partially (weakly) annotated dataset, I choose a completely random subset of PASCAL VOC 2012 with the size of 1000 samples; most experiments are done on this small subset. Furthermore, the partial scribble annotations are from ScribbleSup~\cite{lin2016scribblesup}. Evaluation of all methods is done on the \textit{val} set of PASCAL VOC2012 containing 1449 images, and the mean intersection-over-union (mIoU) is reported.

\textbf{Baseline:} My implementations are based on DeepLab-v2~\cite{chen2017deeplab}. For the encoder part of the model, I consider VGG16 and ResNet50 architectures. The model using VGG16 has an initial learning rate of 0.007, batch size of 8, and adopts the Atrous Spatial Pyramid Pooling (ASPP) schema with 4 branches and rates of $\{6, 12, 18, 24\}$. On the other hand, the model using ResNet50 encoder has an initial learning rate of 0.001, batch size of 6, and adopts the Field-Of-View (FOV) schema with the size of 12. I employ the "poly" learning rate policy just as done in DeepLab-v2, with the little adjustment of using a power value of 1.2 in the scheduler. The optimizer is SGD with a momentum value of 0.9 and a weight decay value of $5\times10^{-4}$.

\textbf{CRF:} for the case of applying dense CRF as a regularization loss, I use the same setup as used by Tang \textit{et el.}~\cite{tang2018regularized}, which uses Gaussian kernels, $w = 2^{-9}$, $\sigma_{rgb} = 15$, $\sigma_{xy} = 100$, and $scale = 0.5$. For the case of applying dense CRF for post-processing, I use $w_1 = 3, w_2 = 4, \sigma_{\gamma} = 1, \sigma_{\alpha} = 67, \sigma_{\beta} = 3$.

\subsection{Experimental Results}
\begin{figure}
\vspace{-1.2cm}
    \hspace{-10mm}
    \includegraphics[width=10.5cm]{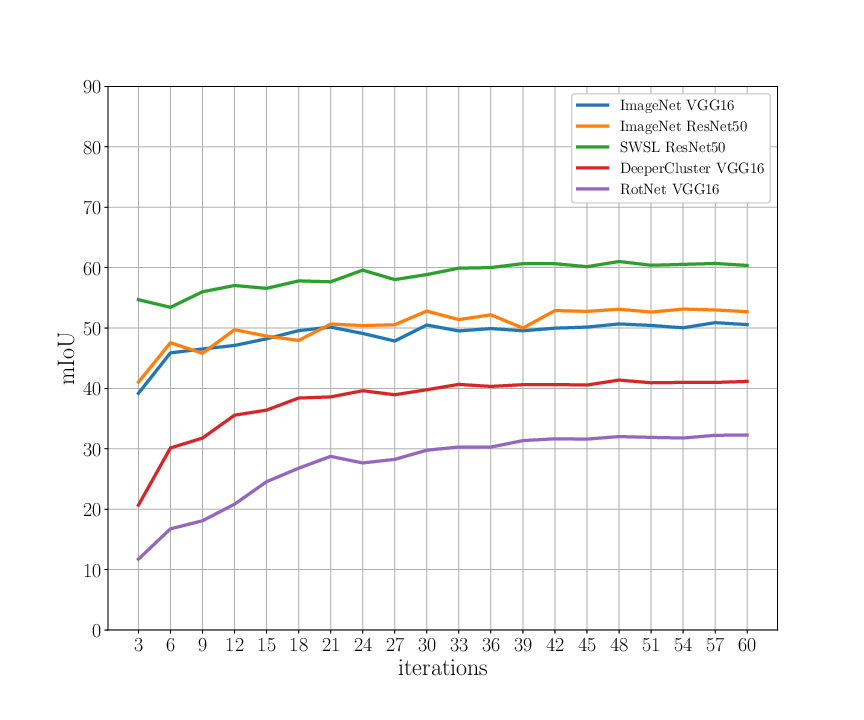}
    \vspace{-0.75cm}
    \caption{mIoU results on PASCAL VOC 2012 \textit{val} set, during training iterations. Case 1: only using pCE}
    \vspace{-0.15cm}
    \label{fig:pce}
\end{figure}

\begin{figure}
\vspace{-0.2cm}
    \hspace{-10mm}
    \includegraphics[width=10.5cm]{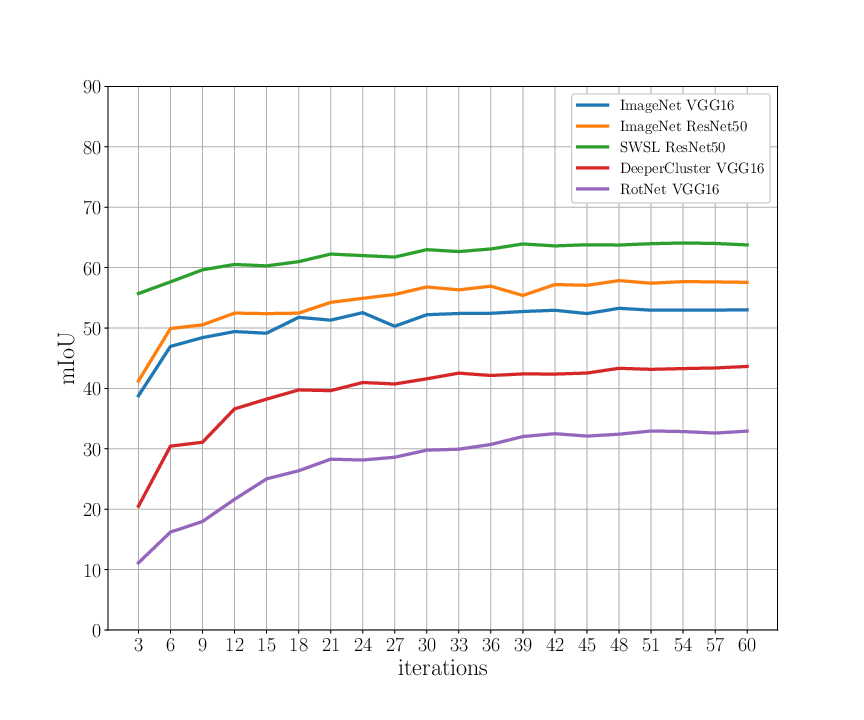}
    \vspace{-0.75cm}
    \caption{mIoU results on PASCAL VOC 2012 \textit{val} set, during training iterations. Case 2: using pCE + dense CRF}
    \vspace{-0.15cm}
    \label{fig:pce_crf}
\end{figure}

\begin{figure}
\vspace{-1.2cm}
    \hspace{-10mm}
    \includegraphics[width=10.5cm]{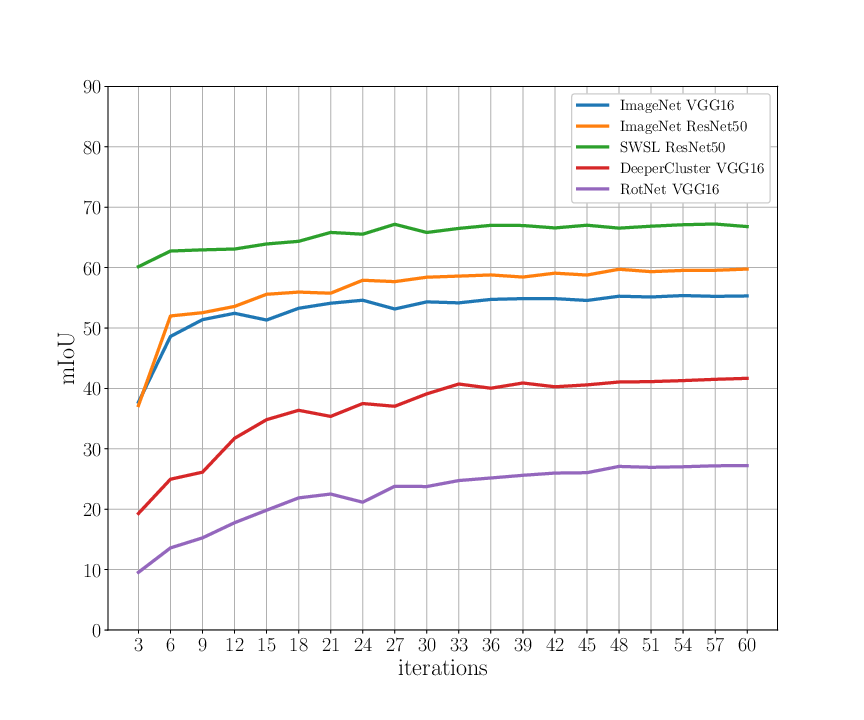}
    \vspace{-0.75cm}
    \caption{mIoU results on PASCAL VOC 2012 \textit{val} set, during training iterations. Case 3: full supervision case.}
    \vspace{-0.15cm}
    \label{fig:full}
\end{figure}

The experiments are conducted in 3 categories: 1) training the network using only partial labels from user interactions (pCE). 2) Training using both pCE and regularized dense CRF loss. 3) The fully annotated case, where the 1000 training samples have pixel-level supervision. For each of these cases, experiments with different pre-training methods are conducted. Table \ref{tab:whole} shows the detailed results. Moreover, as can be seen in this table, the results for both cases of no post-processing and CRF post-processing are reported as well. Figures \ref{fig:pce}
, \ref{fig:pce_crf}, and \ref{fig:full} illustrate the convergence of different setups in each of the 3 categories, respectively.

Figures \ref{fig:pce}, \ref{fig:pce_crf}, \ref{fig:full} and Table \ref{tab:whole} indicate how factors like the CNN architecture, the type of initialization (pre-training), partial labels, and loss regularization can have significant impacts on the results of the segmentation model. Especially, using a ResNet50 SWSL \cite{yalniz2019billion} as the pre-trained network can outperform the case of full-supervision with ImageNet pre-training, even when using only partial labels (pCE). Additionally, when I combine the pCE and dense CRF loss regularization, the model reaches the mIoU performance of 64.1\%, which is 4.3\% better than applying full-supervision with ImageNet pre-training. Replacing the ImageNet pre-training with SWSL pre-training for the full-supervision case enhances the results by 7.4\%. Finally, Table \ref{tab:whole} indicates that using CRF post-processing can be very beneficial as it enhances the performance of every model.

\begin{table*}
\caption{Results of segmentation in 3 scenarios of only pCE, pCE and dense CRF loss regularization, and full supervision. For each case, various models with different per-training approaches are trained and evaluated.}
\vspace{0.1cm}
\centering
\footnotesize
\setlength{\tabcolsep}{0.075cm} 
\begin{tabular}{lcccccc}
 \hline
   ~&  \multicolumn{2}{c}{\bf pCE only} & \multicolumn{2}{c}{\bf pCE + denseCRF} & \multicolumn{2}{c}{\bf Full supervision}\\
   \cmidrule(lr){2-3} \cmidrule(lr){4-5} \cmidrule(lr){6-7} 
       \bf Pre-training method &
       \bf No post-processing &\bf  CRF post-processing &
       \bf No post-processing &\bf  CRF post-processing &
       \bf No post-processing &\bf  CRF post-processing  \\
          \hline
       
       \bf RotNet-VGG16   &  32.3 & 34.8 & 32.9 & 34.0 & 27.2 & 27.7
       \\  

     \bf DeeperCluster-VGG16   &  41.4 & 44.3 & 43.6 & 45.2 & 41.7 & 42.0
       \\
       
     \bf ImageNet-VGG16   &  50.9 & 53.5 & 53.3 & 54.7 & 55.4 & 56.5
       \\
       
     \bf ImageNet-ResNet50   &  53.1 & 56.4 & 57.8 & 60.4 & 59.8 & 61.1
       \\
       
     \bf SWSL-ResNet50   & \bf  61.0 & \bf 64.3 &\bf 64.1 &\bf 66.0 &\bf 67.2 &\bf 68.0
       
       \label{tab:whole}
\end{tabular}

\end{table*}


\subsubsection{Training on the Complete PASCAL VOC 2012 Dataset (10582 Images)}
\begin{figure}
\vspace{-1cm}
    \vspace{12mm}
    \includegraphics[width=8cm]{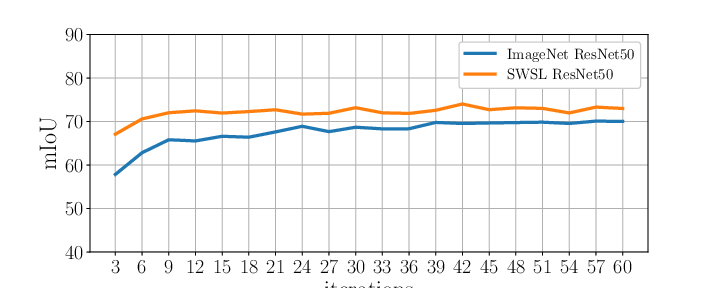}
    \vspace{-1mm}
    \caption{Full-supervision with ImageNet and SWSL pre-training. SWSL pre-training converges much faster and has 3.9\% better mIoU. }
    \vspace{-0.15cm}
    \label{fig:compare}
\end{figure}

Additional to the experiments done on the small dataset (1000 samples), here, I perform and compare the full-supervision training for ImageNet and SWSL pre-training on the complete training data of PASCAL VOC 2012. Figure \ref{fig:compare} illustrates the mIoU results during the training iterations of the models. As can be seen, only a change of pre-training method enhances the mIoU results by 3.9\% for this case.

\begin{figure*}
    \centering
    \includegraphics[width=19cm]{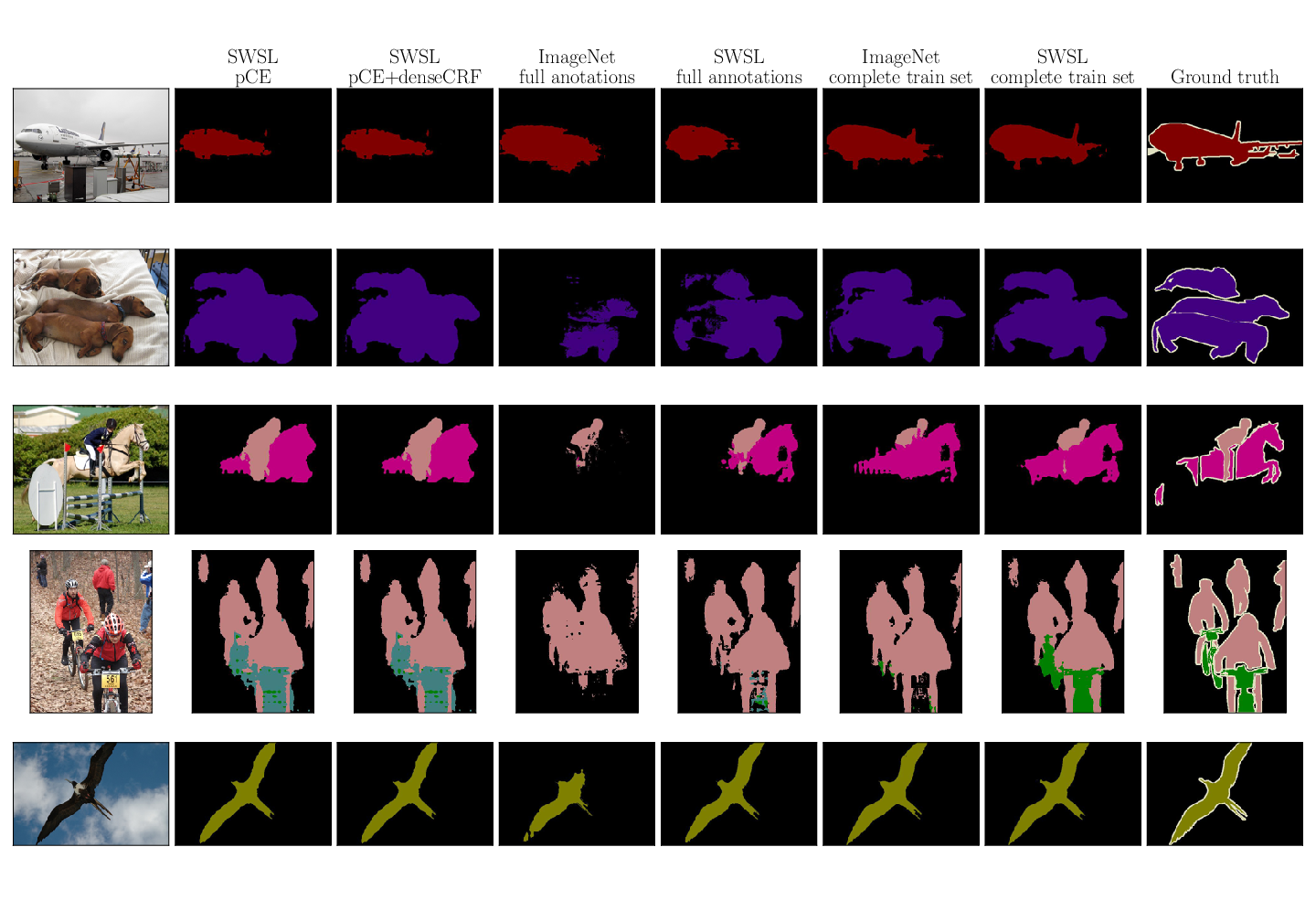}
    \vspace{-1.2cm}
    \caption{Examples of PASCAL VOC 2012. All these results enjoy CRF post-processing. The models using the SWSL pre-training constantly outperform the ones applying ImageNet pre-training.}
    \label{fig:segs}
\end{figure*}

\section{Conclusion}
\label{sec:conclusion}
Experiments show that training DeepLab models on a weakly annotated dataset, using only the partial labels can achieve considerable mIoU results on the PASCAL VOC 2012 \textit{val} set, and when dense CRF loss regularization and CRF post-processing are incorporated to the models, they can achieve near full supervision performances. Most importantly, in this work, the effects of model pre-training are clearly shown, especially, the SWSL pre-training can outperform regular ImageNet pre-training by 7\% and 4\% in the cases of the small dataset and the complete training set, respectively.

\clearpage

{\small
\bibliographystyle{ieee_fullname}
\bibliography{refs}
}

\end{document}